# Machine Learning based prediction of Vanadium Redox Flow Battery temperature rise under different charge-discharge conditions

Anirudh Narayan D[1+] Akshat Johar[1+] Divye Kalra[1+] Bhavya Ardeshna[1+] Ankur Bhattacharjee[1*+]

[1] Department of Electrical and Electronics Engineering, Birla Institute of Technology and Science Pilani, Hyderabad Campus, Telangana-500078, India

[*]*Corresponding Author and Supervisor, Email Id: a.bhattacharjee@hyderabad.bits-pilani.ac.in*
[+]*The authors have equal contributions to this work*

**Abstract**

Accurate prediction of battery temperature rise is very essential for designing an efficient thermal management scheme. In this paper, machine learning (ML) based prediction of Vanadium Redox Flow Battery (VRFB) thermal behavior during charge-discharge operation has been demonstrated for the first time. Considering different currents with a specified electrolyte flow rate, the temperature of a kW scale VRFB system is studied through experiments. Three different ML algorithms; Linear Regression (LR), Support Vector Regression (SVR) and Extreme Gradient Boost (XGBoost) have been used for the prediction work. The training and validation of ML algorithms have been done by the practical dataset of a 1kW 6kWh VRFB storage under 40A, 45A, 50A and 60A charge-discharge currents and 10 L min$^{-1}$ of flow rate. A comparative analysis among the ML algorithms is done in terms of performance metrics such as correlation coefficient ($R^2$), mean absolute error (MAE) and root mean square error (RMSE). It is observed that XGBoost shows the highest accuracy in prediction of around 99%. The ML based prediction results obtained in this work can be very useful for controlling the VRFB temperature rise during operation and act as indicator for further development of an optimized thermal management system.

## 1. Introduction

The abundance of renewable energy sources (RES) such as solar, wind etc. has drawn deeper interest in their rapidly increasing implementation over the past couple of decades. But the challenges lie in their intermittency, weather dependency and thus reliability of power generation. Therefore, storing the energy is necessary to improve the supply reliability at the load end, when the source is not available or inadequate. Among various energy storage solutions considering the faster response time and relatively higher energy density, battery energy storage applications have grown exponentially. Batteries are one of the promising energy storage solutions being adopted in different power system applications. Among the popular battery technologies such as Lead-acid, Li-ion, NiMH etc., vanadium redox flow batteries (VRFB) and other flow batteries have been receiving more priority in the last decade in stationary energy storage applications. Especially in renewable energy applications VRFB is drawing more interest because of its several advantages compared to other conventional batteries. VRFB possesses scalability of its power and energy capacity, longest cycle-life (>20000), deep discharge capacity etc. [1]. The volume of the electrolyte determines the energy capacity of the VRFB, while the stack size determines the power capacity. During the charging/discharging process of the VRFB, the electrolyte acts as a heat carrier for transferring the heat from stack to the tank. Despite having the above-mentioned advantages of VRFB storage, it has a shortcoming of relatively low energy density compared to Li-ion and other non-flow batteries. Therefore, VRFB application is more suited for stationary energy systems. Researchers from across the world have so far reported the work on VRFB stack design, electrode and electrolyte material [2], flow field and control systems for VRFB operation [3-6]. It is to be noted that although during the charge-discharge operation of VRFB, the battery temperature is not so sensitive unlike Li-ion batteries, which undergo thermal runaway and even explosion for temperature rise beyond specified limit. Yet, the VRFB temperature needs to be kept within a certain limit to ensure safe operation. The fast charge-discharge operation is needed in many emergency applications where the VRFB temperature rise may cross the manufacturer-specified limit. By keeping the VRFB temperature within a specified limit of (5°C - 40°C) reduces the chances of side reactions and crossover, and improves electrolyte stability and efficiency [7]. The VRFB temperature can be controlled by optimizing the electrolyte flow rate, as it acts as a coolant as well. Electrolyte flow carries the heat generated inside the stack to the tanks. Hence, to optimize the flow rate for VRFB thermal management, it is necessary to accurately predict the flow battery temperature rise under different charge-discharge conditions. Accurate prediction of battery thermal behavior is an integral part of a battery management system (BMS). Under various operating conditions (charge-discharge rates, state of charge, electrolyte flow rate etc.) the prediction of VRFB temperature rise is necessary to assess its performance and health for on-field applications [8]. So far researchers across the globe have reported their contributions to the prediction of different battery performance parameters (capacity decay, power loss, SOC, SOH etc.). In recent days, the prediction of battery performance parameters using ML algorithms [9-15] is being adopted at a high pace as the data-driven models are free from any accurate or complex mathematical formulation or correlation between the input and output variables. Shen et al. [16] presented a basic Neural etwork based prediction of VRFB system operation and control. Li et al. [17] introduced a U-net based neural network to predict the dynamic of VRFB electrical parameters. In the work reported by Kandasamy et al. [18], Artificial Neural Network (ANN) and ensemble-based ML techniques for predicting the charging/discharging profiles of the stationary battery storage for its optimal performance were discussed. The literature [19] demonstrated the use of a trained Convolution Neural Network (CNN) regression model for predicting the pressure drop while designing the flow fields of redox flow batteries (RFBs). Considering the practical conditions during the charge-discharge process of VRFB, the power loss was predicted by Nawin Ra et al. [20]. In their work, three ML algorithms, namely Linear Regression (LR), Support Vector Regression (SVR) and AdaBoost based ensemble learning were used to predict the power loss of the VRFB under different current levels and electrolyte flow rates. The prediction accuracy comparison was done by the performance metrics such as; $R^2$ (Correlation coefficient), RMSE (Root mean square error) and MAE (Mean Absolute Error).

So far, no ML based prediction of VRFB temperature rise under the impact of different charge-discharge currents and electrolyte flow rate has been reported as per the knowledge of the authors.

For the first time, in this paper the ML algorithm-based prediction of VRFB electrolyte temperature rise for practical charge-discharge profiles and electrolyte flow rate has been performed. Three different ML based predictive algorithms; LR, SVR and XGBoost techniques have been utilized, and their performance accuracy has been analyzed. The ML model training and validation have been done by a practical dataset of a 1kW 6kWh VRFB system operation.

A brief comparison study of the recently published relevant works and the proposed work has been shown in Table 1 to justify its novelty.

**Table 1** Comparison between the proposed work and previous literature regarding ML based prediction of VRFB parameters

| Literature | ML based techniques used | Predicted parameter |
|---|---|---|
| Li et al. [17] | U-Net based Neural Network | Dynamic behavior of VRFB electrical parameters |
| Kandasamy et al. [18] | Neural Networks and Ensemble Techniques | Charge-discharge profiles of stationary battery storage (VRFB, LiFePO4 etc.) |
| Shuaibin Wan et al. [19] | Convolutional neural networks (CNN) | Uniformity factor and pressure-drop of flow Fields for VRFB |
| Nawin Ra et al. [20] | Linear and ensemble learning algorithms | VRFB system power loss |
| Amanda A. Howard et al. [21] | CoKriging (CoPhIK) machine learning method constrained by the zero-dimensional physics-based model | Charge–discharge characteristics curve of VRFB |
| S. Jung et al. [22] | Gaussian Progress regression (GPR) combined with informer model | Long-term capacity fed forecasting for VRFB |
| Proposed work | Linear Regression (LR), Support Vector Regression (SVR), and XGBoost | VRFB stack electrolyte temperature rise under different charging and discharging conditions. |

The proposed work focuses majorly on three objectives:
1. Predicting the VRFB stack electrolyte temperature rise under different charge-discharge conditions using Machine Learning (ML) algorithms.
2. Determining the accuracy of the predictions made by different ML algorithms that utilize widely recognized performance metrics ($R^2$, MAE, RMSE etc.) and validation by comparing with the dataset obtained from a practical VRFB system experimental study.
3. Identifying the suitable ML algorithm for VRFB temperature rise prediction over a range of charge-discharge profiles.

The rest of the chapter is organized as follows; Section 2 presents the overall schematic description of the proposed work; Section 3 describes the VRFB thermal modeling and understanding the thermal behavior; Section 4 provides the mathematical insights of the three Machine Learning techniques used in this work; Section 5.1 comprises of the experimental setup, Section 5.2 represents the experimental results used as input to the ML models, Section 5.3 demonstrates the detailed result and performance analysis; Finally, in Section 6, the conclusion with major contributions of the work have been discussed.

## 2. Overall schematic of the proposed work:

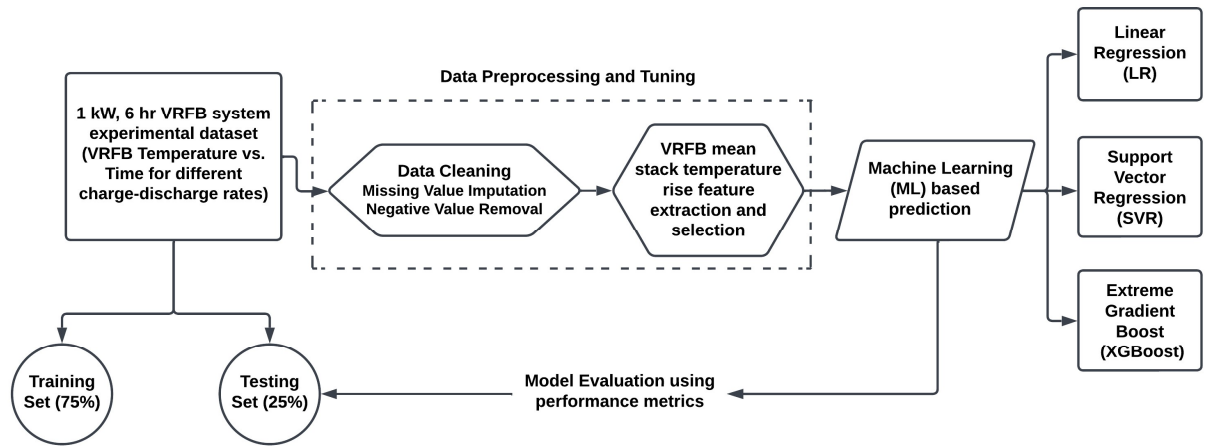

**Fig. 1.** Process flow diagram for ML based VRFB temperature prediction

Fig. 1. shows the entire workflow of the proposed work, i.e. prediction of VRFB electrolyte temperature rise over practical charge-discharge profiles using ML algorithms. The algorithms used are evaluated based on the different metrics such as RMSE, $R^2$ and MAE. Firstly, the temperature data is obtained from the experimental set-up of a 1kW 6kWh VRFB system for different charge-discharge profiles. After collecting the experimental data, the pre-processing and tuning are performed. It is important to perform data pre-processing and tuning to ensure better training dataset and thus improve the prediction accuracy of the ML models. After pre-processing the tuned dataset is now divided into training and testing datasets; 75% and 25% respectively, and the training dataset is fed into three different ML algorithms, i.e. LR, SVR and XGBoost respectively. Finally, performance of the mentioned algorithms has been evaluated based on the different error metrics. RMSE is the root mean square error between the actual/experimental and predicted data. If the prediction performance of the algorithm is poor in nature, then the RMSE appears as high, and thus it can be said that the used algorithm is unreliable for predictive purposes. RMSE indicates the larger errors more significantly than the smaller errors because of its squaring operation. The $R^2$ is the correlation coefficient; degree of relationship between the predicted and experimental results. It ranges from 0 to 1, where '0' is the case for very weak relationship, and '1' is the case for a very strong relationship between predicted and actual/experimental data. A reliable predictive algorithm must have the value of $R^2$ very much near to 1. MAE, as the name suggests, is the mean/average difference between experimental and predicted values for different samples of the dataset. An algorithm that generates strong predicted values possesses a low MAE, and high MAE indicates weak prediction.

## 3. VRFB thermal characteristics model

For effective anticipation and regulation of VRFB electrolyte temperature rise under disparate charge-discharge scenarios, understanding the thermal characteristics model stands as an imperative task. The VRFB storage system consists of a pair of electrolyte tanks, a stack with properly grooved flow channels, and two flow pumps, sensor components, and electrical interface units that connect to a central controller. The flow pumps facilitate the circulation of electrolyte between the tanks and the stack. Unlike conventional batteries, VRFB involves electron transfer at the electrodes within its storage system. The electrolyte in VRFBs contains sulfuric acid with vanadium ions in different oxidation states. In the realm of low temperatures (below 5°C), the negative electrolyte experiences the precipitation of $V^{2+}/V^{3+}$, while elevated temperatures (exceeding 45°C) lead to the precipitation of $V^{5+}$ in the positive electrolyte. This precipitation phenomenon has the potential to obstruct electrolyte channels, degrade the membrane, and consequently degrade battery performance.

For comprehensive realization of the thermal characteristics of VRFB storage across diverse operational scenarios, multiple endeavors have been undertaken to model its thermal behavior. Zheng et al. [23] devised a three-dimensional VRFB model, whereas Al-Fetlawi et al. [24] developed a non-isothermal model specifically addressing temperature variations within the VRFB stack under varying operational conditions. Tang et al. [25, 26] and Yan et al. [27] explored thermal modeling alongside the influence of self-discharge on VRFB thermal behavior. Zhang et al. [28] delved into the impact of operational temperature on VRFB system performance and capacity decline. Trovò et al. [29] showcased a kW scale VRFB thermal management system for standby scenarios.

The thermodynamic equations [30, 31] of VRFB system during charging and discharging are represented by Eq. (1-4) as below;

$$C_P \rho V_S \frac{dT_S}{dt} = Q_+ C_P \rho (T_+ - T_S) + Q_- C_P \rho (T_- - T_S) + I_C^2 R_C + I_C T_S \frac{dE}{dT} \qquad (1)$$

$$C_P \rho V_S \frac{dT_S}{dt} = Q_+ C_P \rho (T_+ - T_S) + Q_- C_P \rho (T_- - T_S) + I_D^2 R_D + I_D T_S \frac{dE}{dT} \quad (2)$$

$$C_P \rho V_+ \frac{dT_+}{dt} = Q_+ C_P \rho (T_S - T_+) + U_+ A_+ (T_a - T_+) \quad (3)$$

$$C_P \rho V_- \frac{dT_-}{dt} = Q_- C_P \rho (T_S - T_-) + U_- A_- (T_a - T_-) \quad (4)$$

Where,
$C_p$ = Specific heat of the electrolyte (J g⁻¹ K⁻¹)
$\rho$ = Density of electrolyte (g m⁻³)
$V_-$ = Volume of negative electrolyte tank (L)
$V_+$ = Volume of positive electrolyte tank (L)
$V_S$ = Volume of electrolyte inside the battery stack (L)
$T_+$ = Temperature of positive electrolyte in the tank (°C)
$T_-$ = Temperature of negative electrolyte in the tank (°C)
$T_S$ = Temperature of electrolyte in the stack (°C)
$T_a$ = Ambient temperature (°C)
$Q_+$ = Outlet flow rate of positive electrolyte from the tank (L min⁻¹)
$Q_-$ = Outlet flow rate of negative electrolyte from the tank (L min⁻¹)
$U_+$ = Overall heat transfer coefficient of the tank on the positive electrolyte side (W m⁻² K⁻¹)
$U_-$ = Overall heat transfer coefficient of the tank on the negative electrolyte side (W m⁻² K⁻¹)
$I_C$ = Charging current (A)
$I_D$ = Discharging current (A)
$A_+$ = Surface area of the positive side of the tank (m²)
$A_-$ = Surface area of the negative side of the tank (m²)
$R_C$ = Overall stack resistance during charging (Ω)
$R_D$ = Overall stack resistance during discharging (Ω)

The heat transfer rate caused by the electrolyte flow between stack and tanks is indicated by the initial two terms on the right-hand side of Eq. (1) and (2), during charging and discharging respectively. The heat that results from ohmic resistance during the charging and discharging process are represented by the third term ($I_C^2 R_C$) and ($I_D^2 R_D$). In the last term of Eq. (1) and (2), the model also includes a specific term for reversible entropic heat in each case of charging and discharging; ($I_C T_S \frac{dE}{dT}$), and ($I_D T_S \frac{dE}{dT}$) respectively. This 'E' is the open circuit voltage of VRFB stack and it is represented by Eq. (5) 'Nernst Equation' as follows;

$$E = n \times \left\{ E_0 + \frac{2RT}{F} \ln\left(\frac{SOC}{1-SOC}\right) - I_d R_{sd} \right\} \quad (5)$$

Where,
$E$ = Open circuit voltage of VRFB stack (V)
$E_0$ = The equilibrium potential (at 50% SOC)
$R$ = Universal gas constant (8.3144 $J\ K^{-1} mol^{-1}$)
$T$ = Ambient temperature (K)

n = No. of series cells in VRFB stack
$I_d$ = Diffusion current (μA)
$R_{sd}$ = Self discharge potential drop equivalent resistance (MΩ)
In this work, the impact of VRFB self-discharge on the stack electrolyte temperature has not been considered because of its negligibly small value during idle operating condition, compared to the heat generation due to the charge-discharge current and electrolyte flow rate [26].
Here, both $Q_+$ and $Q_-$ have been considered as equal because of the uniform speed maintained for the two pumps. The heat generated inside the stack is carried out by the electrolyte to the two tanks. As the electrolyte flow rate maintained by the two pumps are equal, the heat carried by the electrolyte from the stack to the two tanks are also equal and occur in real-time. Hence, the temperature of both the positive and negative electrolyte tanks ($T_+$ and $T_-$) are assumed as equal at a time [25].
Based on the above-mentioned logical assumptions and considerations, the Eq. (1) and (2) have been further simplified and expressed by Eq. (6) and Eq. (7);

$$T_{s\_C} = \frac{\frac{Q}{V_s} t_C (T_{+/-})}{C_p \rho \left(V_s + 2Q t_C - \frac{I_C \frac{dE}{dT}}{c_p \rho}\right)} + \frac{I_C^2 R_C t_C}{C_p \rho \left(V_s + 2Q t_C - \frac{I_C \frac{dE}{dT}}{c_p \rho}\right)} \quad (6)$$

$$T_{s\_D} = \frac{\frac{Q}{V_s} t_D (T_{+/-})}{C_p \rho \left(V_s + 2Q t_D - \frac{I_D \frac{dE}{dT}}{c_p \rho}\right)} + \frac{I_D^2 R_D t_D}{C_p \rho \left(V_s + 2Q t_D - \frac{I_D \frac{dE}{dT}}{c_p \rho}\right)} \quad (7)$$

Where,
$T_+ = T_- = T_{+/-}$ = Temperature of electrolyte in the positive and negative electrolyte tank
$Q = Q_+ = Q_-$ (L min$^{-1}$)
$T_{s\_C}$ = VRFB stack electrolyte temperature during charging (°C)
$T_{s\_D}$ = VRFB stack electrolyte temperature during discharging (°C)
$t_C$ = Charging duration (Hour)
$t_D$ = Discharging duration (Hour)

In this paper, considering the long period of experiment runtime of about 4-5 hours for the kW scale commercial VRFB storage system, a constant optimal flow rate has been selected as a case study. The VRFB stack electrolyte temperature rise with the change in stack current during charging and discharging has been predicted by ML algorithms. In the experimental case study, this flow rate has been kept at a higher level but within the manufacturer-specified limit, so that it can primarily serve the purpose of restricting rapid temperature rise during charge-discharge process, by faster transfer of heat generated inside the VRFB stack to the tanks. Fig. 2 represents the generalized block diagram of VRFB thermal characteristics model.

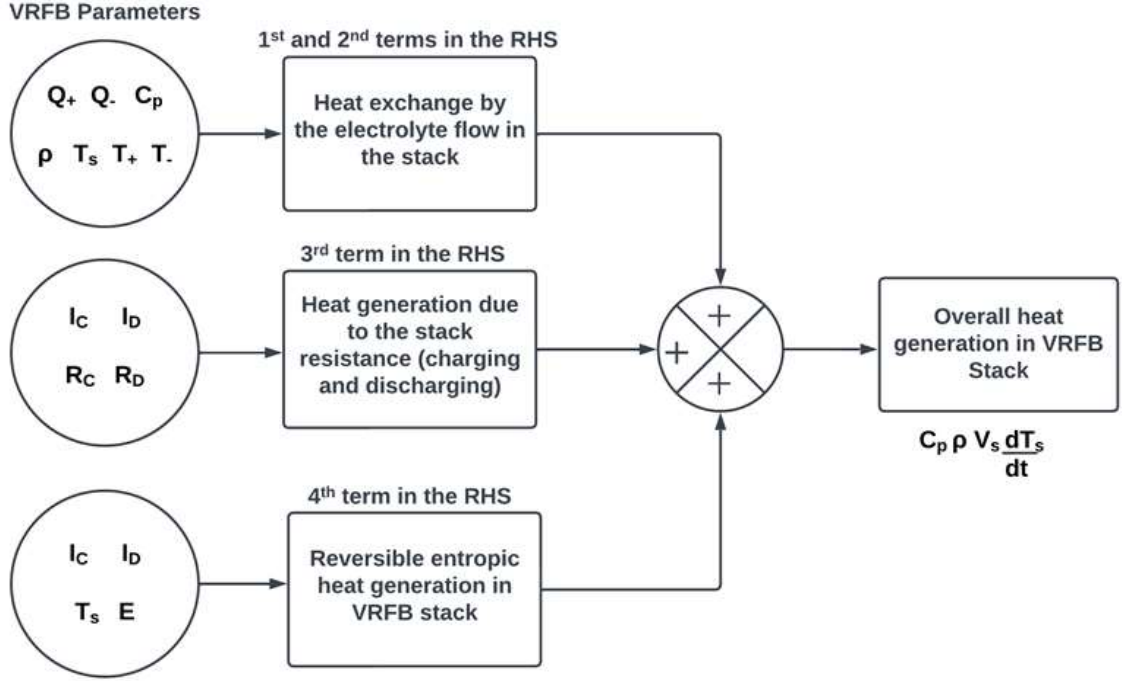

**Fig. 2** Generic block diagram representation of the VRFB thermal characteristics model for charge-discharge conditions

## 4. Machine Learning Models for the proposed work

Considering the data-driven model for the prediction of VRFB stack electrolyte temperature rise during charge-discharge, three different ML algorithms have been applied, and their performance has been analyzed based on the error metrics.

Linear Regression (LR) [32-35] is a regression technique utilized to understand the interrelationship that exists between the dependent and independent variables. The main goal of linear regression is to find the best-fitting straight line (or hyperplane in higher dimensions) that effectively describes the connection between these variables. Once the linear equation is ascertained, it can be used to forecast the dependent variable values when provided by new independent variable values. The functioning of LR involves minimizing the sum of squared deviations between forecasted and observed values of the dependent variable. LR can be categorized as univariate or multivariate depending on the count of independent variables involved. In the context of this study, the univariate dataset is utilized to predict the temperature of the VRFB stack electrolyte during operation. Eq. (8) represents the loss function ($J$) for a single feature LR model.

$$J = \frac{1}{n}\sum_{i=1}^{n} L(\overline{y_i} - y_i)^2 \qquad (8)$$

Where,
$y$ = Experimental VRFB stack electrolyte temperature
$\overline{y}$ = Anticipated temperature prediction
$n$ = Number of data points

$L$ = Loss function

Support Vector Regression (SVR) [36,37] is a regression technique that is based on the lines of Support Vector Machines (SVM) used for prediction of values that are continuous. The main objective is to construct a hyperplane that fits the training data best, with some tolerance. Eq. (9) represents the objective function of SVR algorithm.

$$M = min \frac{1}{2}|w|^2 + C \sum_{i=1}^{n} |\xi_i| \qquad (9)$$

Subject to constraints 'Z', defined as;

$$Z = |y_i - w_i x_i| \leq \varepsilon + |\xi_i| \qquad (10)$$

Where,

$M$ = Objective function of SVR

$|w|$ = Magnitude of the weight vector

$C$ = Tolerance-adjusting hyper-parameter for points outside 'ε'

$n$ = Number of data points in the 'time vs. VRFB electrolyte temperature' dataset

$\varepsilon$ = Maximum error

$\xi$ = Deviation from the maximum error 'ε'

$x$ = Experimental VRFB stack electrolyte temperature

$y$ = Predicted VRFB stack electrolyte temperature

For the SVR technique, the well-known radial basis function (RBF) kernel has been used, due to its ability to suitably deal with problems where the underlying data distribution is non-linear. The RBF kernel results in smooth and continuous predictions. Its high flexibility makes it desirable in many regression problems. Regarding the execution of SVR algorithm, it should be noted that in Eq. (10) 'ε' signifies the maximum error, that is the tolerance in deviation of predicted value from the boundary of the ε-insensitive zone. For each value of 'ε' the SVR model has been trained using the training dataset. After running multiple simulations for the dataset used in the case studies with different levels of VRFB charge-discharge current, the value of 'ε' is in the range of 0.04 - 0.1 for 40A - 60A case studies respectively. The maximum permissible error value of 0.1 has been observed in the case of the 50A discharging dataset.

Tree boosting [38,39] is a widespread and powerful machine learning technique. In this paper, a tree boosting system algorithm called Extreme Gradient Boosting (XGBoost) for short is used. XGBoost is a highly proficient implementation of the gradient boosting algorithm. In this work, XGBoost ML algorithm has been chosen because of its capability of handling large datasets and tuning with missing datapoints with the help of its hyper-parameter tuning feature.

The objective function of the XGBoost algorithm (loss function and regularization) at iteration that needs to be minimized is expressed by Eq. (11).

$$L(y, p) = \sum_{i=1}^{n} L(y_i, p_i) + \frac{1}{2}\lambda O_V^2 \qquad (11)$$

Where,

$L(y_i, p_i)$ = Normal Regression loss between predicted and experimental temperature

$\lambda$ = Hyperparameter for L2 regularization on leaf weights

$O_V$ = Similarity weight

Eq. (11) consists of two terms. The first term is the loss function, and the second is the regularization term.

The goal is to minimize the loss function. The loss function of XGBoost is obtained by using second-order Taylor approximation. For both regression and classification tasks, the loss function can be approximated by Eq. (12).

$$L(y, p_i^0 + O_V) = L(y, p_i) + [\frac{d}{dp_i}L(y, p_i)]O_V + \frac{1}{2}[\frac{d^2}{dp_i^2}L(y, p_i)]O_V^2 \qquad (12)$$

The value of '$O_V$' after minimizing the loss function is presented by,

$$O_V = \frac{-(g_1+g_2+\cdots+g_n)}{(h_1+h_2+\cdots+h_n+\lambda)} \qquad (13)$$

$$O_V = \frac{(y_1-p_1)+(y_2-p_2)+\cdots+(y_n-n)}{(1+1+\cdots+1+\lambda)} \qquad (14)$$

$$O_V = \frac{Sum\ of\ residuals}{Number\ of\ residuals+\lambda} \qquad (15)$$

Where,

$g$ = First derivative related to gradient descent

$h$ = Second derivative related to Hessian

While solving the values of '$g$' and '$h$', it turns out that all the values of '$g$' become the difference between the predicted value and actual value, and '$h$' becomes equal to 1. Here, Eq. (15) represents the output for XGBoost in regression.

In this paper, the residuals are defined as the difference between a particular stack electrolyte temperature value for the specified charging-discharging current rates of VRFB, and the mean temperature for the same.

## 5. Results and Discussion

### 5.1 Experimental Setup

In order to validate the performance of ML algorithms in predicting VRFB stack electrolyte temperature rise during charge-discharge operations, a practical 1kW 6kWh VRFB system has been used for experimentation. Fig. 3 shows the complete setup of the VRFB system [40], along with its pressure and temperature sensor interface and other electronic subsystems which are required for smooth operation. As mentioned earlier, the heat generated inside the VRFB stack during operation is carried by the flow of electrolyte from the VRFB stack to the tanks. At a considerably higher flow rate within the specified range of the VRFB manufacturer, the heat transfer through the electrolyte from the stack to tanks happens even faster. As a result, the temperature difference between the VRFB stack and the tank becomes very small. This fact has also been reported in some existing literature on VRFB thermal behaviour assessment. In one such work [7], it is observed that during operation the VRFB stack and tank temperature values are very close (~ 0.5°C) to each other, at a higher flow rate. In this work the temperature sensor has been installed in the tank to avoid the installation constraints inside the stack for a commercial VRFB system, and the tank temperature rise profile very closely indicates the stack electrolyte temperature rise [41] at a higher electrolyte flow rate. A Programmable Logic Controller (PLC) unit is installed for centralized monitoring and control of VRFB operation. The specifications for the VRFB system chosen in this work are given in Table 2.

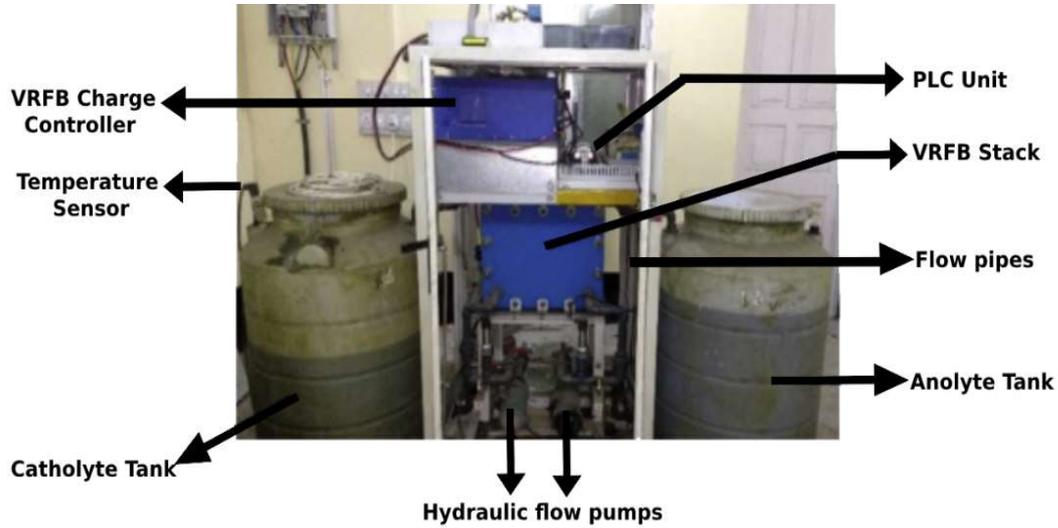

**Fig. 3** Experimental setup of 1 kW 6 kWh VRFB system

**Table 2** Specifications of VRFB storage system

| VRFB System Parameters | Rating |
|---|---|
| Power capacity | 1 kW |
| Energy Capacity | 6kWh |
| Rated stack terminal current | 60 A |
| Number of series cells in stack | 20 |
| Stack terminal voltage range | (20-32) V |
| Dimension of individual electrode ($L_{felt} \times W_{felt} \times D_{felt}$) | 25 cm × 25 cm × 0.3 cm |
| Range of electrolyte flow rate | (1-18) L min$^{-1}$ |
| Current density | 0.096 A cm$^{-2}$ |
| Electrolyte concentration | 1.2 mol L$^{-1}$ |
| Ambient operating temperature range | (15 – 35)°C |

## 5.2 Experimental results: Input to the Machine Learning models

The dataset of VRFB electrolyte temperature obtained from the experiments for four different charge-discharge currents at an average optimal flow rate of 10L min$^{-1}$ for the 1kW 6kWh VRFB system, at an ambient temperature of 30°C is shown in Fig. 4a and 4b. As observed, when the current increases the charging and discharging time reduces. It is to be noted that in fast charging/discharging cases, a large amount of heat is generated inside the VRFB stack which has a strong impact on the battery performance [7] e.g; affecting the thermodynamics and kinetics of electrochemical reactions, physicochemical properties of the key components of the battery, salt precipitation, etc. Hence, it is important to predict the VRFB stack electrolyte temperature rise for a wide range of charge-discharge operations, which will further be useful for designing an optimal thermal management system [29,41,42], considering various operating temperature conditions [43] and practical applications such as EV charging stations [44]. As mentioned above, in this paper, considering a long period of experiment runtime of about 4-5 hours for the kW scale VRFB storage system, the constant optimal flow rate has been selected (10 L min$^{-1}$ ~ 170 ml/sec) as a case study for predicting the electrolyte temperature rise in VRFB with the change in stack current during charging and discharging. This flow rate has been kept at a higher level but within the manufacturer-specified flow rate range (1-18 L min$^{-1}$) so that it can primarily serve the purpose of restricting rapid temperature rise during the charge-discharge process, by faster transfer of heat generated inside the VRFB stack to the tanks.

Different ML models have been used for predicting the VRFB thermal behavior under the above-mentioned practical operating conditions.

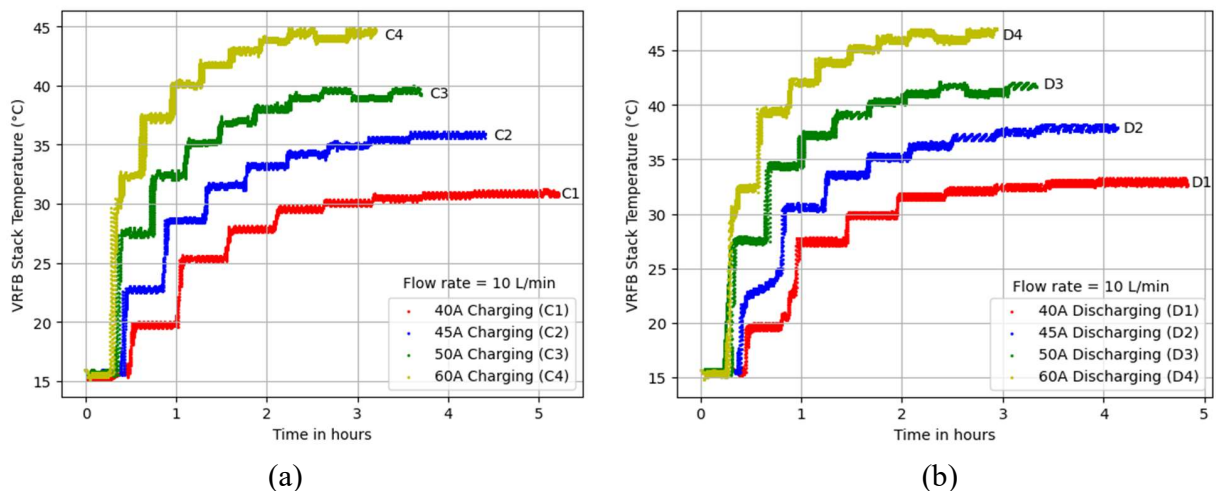

(a)          (b)

**Fig. 4.** VRFB stack electrolyte temperature rise experimental results a) charging profile (b) discharging profile

Table 3 displays the data for the mean and the maximum temperature for different values of stack current (40A, 45A, 50A, and 60A) for both charging and discharging experiments on 1kW 6kWh VRFB. The mean/average and the maximum temperature indicate the input dataset to the ML models. These experimental data have been used to train the ML models. It is to be noted that here the dependent variable is 'stack electrolyte temperature', and the independent variable is the 'time' for the ML algorithms.

**Table 3** Average and maximum VRFB stack electrolyte temperature for different charge-discharge currents of a 1kW 6h VRFB system as input to the ML models

| Stack Current (A) | Mode of operation | Average VRFB stack electrolyte temperature (°C) | Maximum VRFB stack electrolyte temperature (°C) |
|---|---|---|---|
| 40 | Charging | 27.743 | 31.2529 |
|  | Discharging | 25.719 | 33.3343 |
| 45 | Charging | 31.302 | 35.5717 |
|  | Discharging | 33.222 | 38.2843 |
| 50 | Charging | 33.926 | 40.0213 |
|  | Discharging | 36.236 | 42.2443 |
| 60 | Charging | 37.307 | 44.8636 |
|  | Discharging | 39.422 | 47.0195 |

### 5.3 Prediction results and performance validation

A 1kW 6kWh VRFB charge-discharge experimental data set has been used to predict its stack electrolyte temperature rise under different current levels using ML algorithms. The Google Colab platform has been used to train and validate all the ML models. A set of 7095 data samples has been used and the process is executed on PC with a configuration of 16GB RAM and an Intel i7 processor. The performance accuracy of the ML models is computed based on the error metrics such as $R^2$, MAE, and RMSE, obtained from the implemented ML model results. As mentioned in Section 2, the VRFB stack electrolyte temperature experimental dataset is split into the ratio of 75:25 where 75% (5321) of the data is used for training, and the remaining 25% (1774) for testing. The VRFB stack electrolyte temperature prediction is carried out using three ML models namely LR, SVR, and XGBoost as mentioned in Section 4. Data pre-processing is an important step that needs to be done before the implementation of the ML models on the dataset as mentioned in Section 2. Among the chosen three ML models, it is observed that the XGBoost algorithm provides the largest $R^2$ close to 1 for predicting the VRFB stack electrolyte temperature rise under different stack current profiles. As shown in Table 4 for a case study, considering the predicted values of the VRFB stack electrolyte temperature with flow rate 10 L min$^{-1}$ under 40A discharging current the XGBoost technique provides an $R^2$ of 0.99, whereas SVR and LR output scores of 0.96 and 0.69 respectively. The other performance metrics such as RMSE and MAE for XGBoost are 0.32 and 0.24 respectively, whereas the RMSE and MAE for SVR are 0.81 and 0.52 respectively, and those for LR are 2.58 and 2.12 respectively. A similar trend is observed in other discharging and charging conditions, as shown in Table 4. The XGBoost algorithm shows better performance with the highest accuracy in the prediction of VRFB stack electrolyte temperature compared to the other two ML algorithms: LR and SVR.

**Table 4** Prediction performance comparison of ML models

| VRFB stack current (A) | Mode of operation | Error Parameters | | | | | | | | |
| --- | --- | --- | --- | --- | --- | --- | --- | --- | --- | --- |
| | | LR | | | SVR | | | XG Boost | | |
| | | $R^2$ | MAE | RMSE | $R^2$ | MAE | RMSE | $R^2$ | MAE | RMSE |
| 40 | Charging | 0.70 | 1.87 | 2.21 | 0.97 | 0.49 | 0.78 | 0.99 | 0.22 | 0.32 |
| | Discharging | 0.69 | 2.12 | 2.58 | 0.96 | 0.52 | 0.81 | 0.99 | 0.24 | 0.32 |
| 45 | Charging | 0.75 | 2.14 | 2.55 | 0.96 | 0.66 | 1.11 | 0.99 | 0.30 | 0.60 |
| | Discharging | 0.72 | 2.49 | 2.96 | 0.97 | 0.60 | 0.97 | 0.99 | 0.29 | 0.52 |
| 50 | Charging | 0.71 | 2.54 | 3.30 | 0.95 | 0.83 | 1.38 | 0.97 | 0.52 | 1.10 |
| | Discharging | 0.70 | 2.84 | 3.48 | 0.94 | 0.96 | 1.54 | 0.98 | 0.46 | 0.96 |
| 60 | Charging | 0.70 | 4.08 | 5.04 | 0.96 | 0.54 | 0.84 | 0.98 | 0.55 | 1.16 |
| | Discharging | 0.64 | 3.48 | 4.60 | 0.95 | 1.23 | 2.04 | 0.98 | 0.51 | 1.16 |

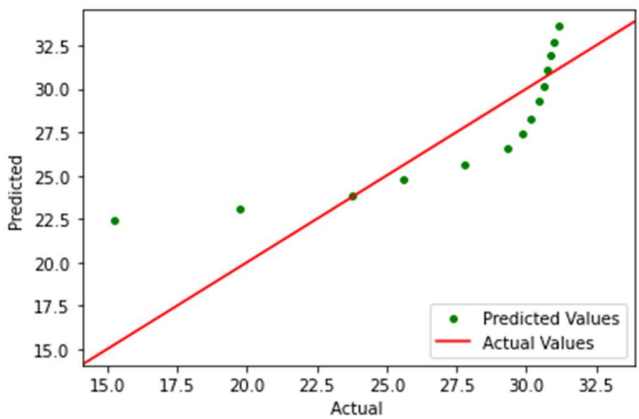

(a) LR - Charging

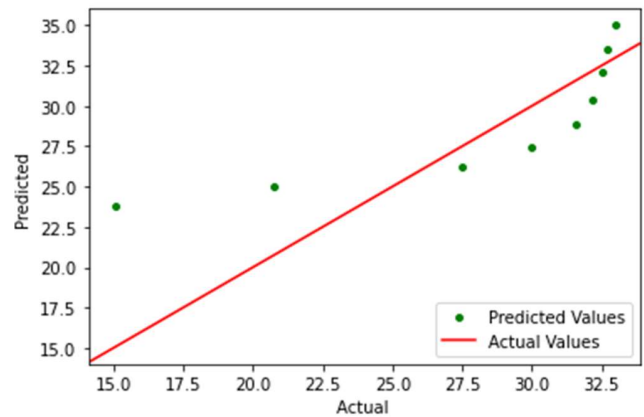

(b) LR – Discharging

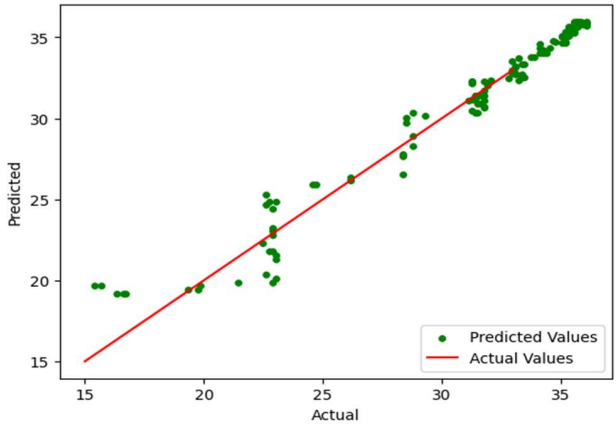

(c) Linear SVR - Charging

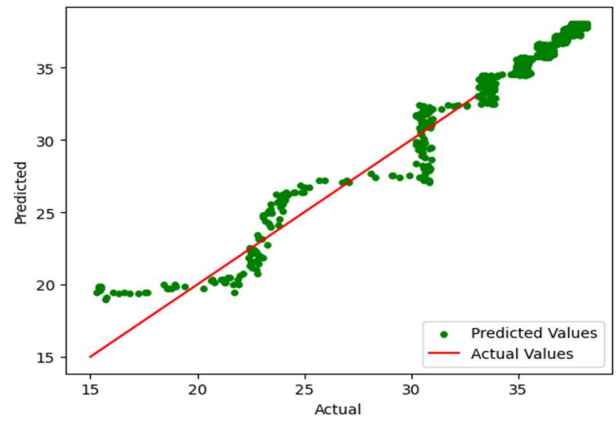

(d) Linear SVR - Discharging

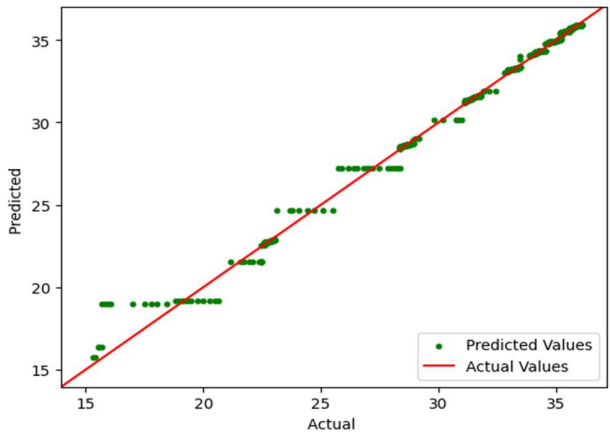

(e) XGBoost – Charging

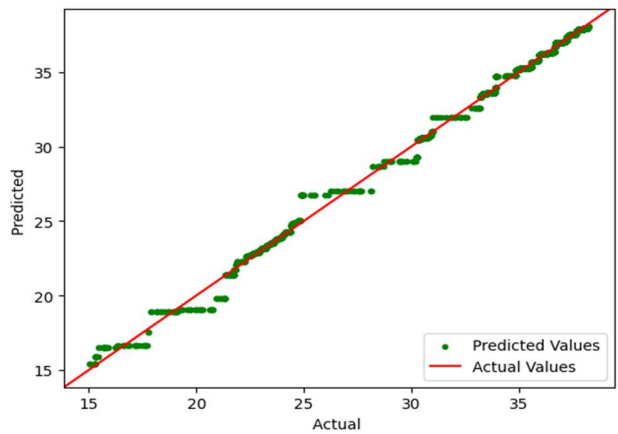

(f) XGBoost – Discharging

**Fig. 5.** Actual vs. Predicted VRFB stack electrolyte temperature for 45A charge-discharge profile at a flow rate of 10L min$^{-1}$ using ML techniques namely: (a) & (b) LR; (c) & (d) SVR; (e) & (f) XGBoost

Fig. 5 describes the proximity of actual and predicted values for all the ML algorithms under 40A, 10 L min$^{-1}$ electrolyte stack current flow rate as one of the case studies. The $R^2$ of XGBoost for 45A current under both charging and discharging conditions was found to be 0.99, as is evident from the graphs in Fig. 5 (e)-(f). The XGBoost graph exhibits minimal deviation from the experimental temperature values, validating its high $R^2$. However, in Fig. 5 (a)-(d), an increase in the deviation of predicted values from the experimental values is observed where LR and Linear SVR algorithms are applied.

This inference is endorsed by the data shown in Table 5, where the relative percentage error is calculated. The range of the relative percentage errors between the experimental values of VRFB stack electrolyte temperature and the XGBoost-based prediction values for all the current levels lies between 0.0083% and 0.1261%. The consistency of such a low percentage error implies that the XGBoost-based prediction is highly accurate.

**Table 5** Experimental vs. Predicted values obtained using the XGBoost algorithm

| VRFB stack current (A) | Mode of Operation | Mean VRFB stack electrolyte temperature rise (°C) | | Relative percentage error = (ER-PR)/ER*100 (%) |
|---|---|---|---|---|
| | | Experimental result (ER) | XGBoost Prediction result (PR) | |
| 40 | Charging | 27.8080 | 27.8224 | 0.0515 |
| | Discharging | 29.7845 | 29.7870 | 0.0083 |
| 45 | Charging | 31.3459 | 31.3854 | 0.1261 |
| | Discharging | 33.3943 | 33.3867 | 0.0228 |
| 50 | Charging | 34.4889 | 34.4961 | 0.0208 |
| | Discharging | 36.4302 | 36.4353 | 0.0140 |
| 60 | Charging | 37.4080 | 37.4257 | 0.0472 |
| | Discharging | 39.6062 | 39.5795 | 0.0674 |

## 6. Conclusion

For the first time ML based prediction of VRFB stack electrolyte temperature rise under various charging-discharging conditions is demonstrated in this work. Considering different charge-discharge current levels with constant optimized electrolyte flow rate within the manufacturer's specified limit, the stack electrolyte temperature rise of a kW scale VRFB system has been studied through experiments. As mentioned earlier at a higher flow rate, the tank electrolyte temperature variation indicates the stack electrolyte temperature variation as the electrolyte carries the heat from stack to tanks and almost in real-time. For predicting the VRFB stack electrolyte temperature, the ML algorithms have been trained and their performance has been validated by a practical dataset of a 1kW 6kWh VRFB storage system under 40A, 45A, 50A, and 60A charge-discharge currents and a constant optimized flow rate of 10 L min$^{-1}$ which is at the higher side of the specified range. The comparative analysis among the three ML algorithms; Linear Regression, Support Vector Regression, and XGBoost has been done in terms of performance metrics such as correlation coefficient ($R^2$), mean absolute error (MAE), and root mean square error (RMSE). It is observed that XGBoost shows the highest prediction accuracy. The XGBoost model exhibits the best prediction accuracy of $R^2$ = 0.99, with the least error parameter of around MAE = 0.24 and RMSE = 0.32 for 40A discharging as a test case. In addition to this, the said ML model accurately predicts the VRFB stack electrolyte temperature rise for the other charge-discharge current levels. The experimental and predicted values of temperature using the XGBoost algorithm have been compared and the relative percentage error is found to be in a range between 0.0083% and 0.1261%. The very low values of error percentages imply that the prediction of temperature using the XGBoost algorithm has the highest accuracy. Therefore, the suggested ML based model for prediction of VRFB stack electrolyte temperature rise during various charge-discharge scenarios claims to be very useful for optimizing the thermal management scheme, thus making the VRFB system operation more efficient on field. The proposed work is a generalized one and can be applicable to large scale VRFB storage systems as well.

**Acknowledgements**

**References**

[1] X.Z. Yuan, C. Song, A. Platt, N. Zhao, H. Wang, H. Li, K. Fatih, D. Jang, A review of all-vanadium redox flow battery durability: Degradation mechanisms and mitigation strategies, Int J Energy Res. 43 (2019). https://doi.org/10.1002/er.4607.

[2] M. Gencten, Y. Sahin, A critical review on progress of the electrode materials of vanadium redox flow battery, Int J Energy Res. 44 (2020). https://doi.org/10.1002/er.5487.

[3] Z. Huang, A. Mu, Research and analysis of performance improvement of vanadium redox flow battery in microgrid: A technology review, Int J Energy Res. 45 (2021). https://doi.org/10.1002/er.6716.

[4] N. Ra, A. Bhattacharjee, An Extensive Study and Analysis of System Modeling and Interfacing of Vanadium Redox Flow Battery, Energy Technology. 9 (2021). https://doi.org/10.1002/ente.202000708.

[5] T. Zou, L. Luo, Y. Liao, P. Wang, J. Zhang, L. Yu, Study on operating conditions of household vanadium redox flow battery energy storage system, J Energy Storage. 46 (2022). https://doi.org/10.1016/j.est.2021.103859.

[6] P. Skupin, S.R. Ambati, Nonlinear model predictive control of vanadium redox flow battery, J Energy Storage. 62 (2023) 106905. https://doi.org/10.1016/j.est.2023.106905.

[7]     J. Ren, Y. Li, Z. Wang, J. Sun, Q. Yue, X. Fan, T. Zhao, Thermal issues of vanadium redox flow batteries, Int J Heat Mass Transf. 203 (2023). https://doi.org/10.1016/j.ijheatmasstransfer.2022.123818.

[8]     A. Trovò, A. Saccardo, M. Giomo, M. Guarnieri, Thermal modeling of industrial-scale vanadium redox flow batteries in high-current operations, J. Power Sources 424 (2019) 204–214, https://doi.org/10.1016/j.jpowsour.2019.03.080

[9]     D. Roman, S. Saxena, V. Robu, M. Pecht, D. Flynn, Machine learning pipeline for battery state-of-health estimation, Nat Mach Intell. 3 (2021). https://doi.org/10.1038/s42256-021-00312-3.

[10]    Z. Fei, F. Yang, K.L. Tsui, L. Li, Z. Zhang, Early prediction of battery lifetime via a machine learning based framework, Energy. 225 (2021). https://doi.org/10.1016/j.energy.2021.120205.

[11]    M.F. Ng, J. Zhao, Q. Yan, G.J. Conduit, Z.W. Seh, Predicting the state of charge and health of batteries using data-driven machine learning, Nat Mach Intell. 2 (2020). https://doi.org/10.1038/s42256-020-0156-7.

[12]    A. Dineva, B. Csomós, S. Kocsis Sz., I. Vajda, Investigation of the performance of direct forecasting strategy using machine learning in State-of-Charge prediction of Li-ion batteries exposed to dynamic loads, J Energy Storage. 36 (2021). https://doi.org/10.1016/j.est.2021.102351.

[13]    Z. Xu, J. Wang, Q. Fan, P.D. Lund, J. Hong, Improving the state of charge estimation of reused lithium-ion batteries by abating hysteresis using machine learning technique, J Energy Storage. 32 (2020). https://doi.org/10.1016/j.est.2020.101678.

[14]    J.K. Thomas, H.R. Crasta, K. Kausthubha, C. Gowda, A. Rao, Battery monitoring system using machine learning, J Energy Storage. 40 (2021). https://doi.org/10.1016/j.est.2021.102741.

[15]    J. Bao, V. Murugesan, C.J. Kamp, Y. Shao, L. Yan, W. Wang, Machine Learning Coupled Multi-Scale Modeling for Redox Flow Batteries, Adv Theory Simul. 3 (2020). https://doi.org/10.1002/adts.201900167.

[16]    H.F. Shen, X.J. Zhu, M. Shao, H.F. Cao, Neural network predictive control for vanadium redox flow battery, J Appl Math. 2013 (2013). https://doi.org/10.1155/2013/538237.

[17]   R. Li, B. Xiong, S. Zhang, X. Zhang, Y. Li, H. Iu, T. Fernando, A novel U-Net based data-driven vanadium redox flow battery modeling approach, Electrochim Acta. 444 (2023) 141998. https://doi.org/10.1016/j.electacta.2023.141998.

[18]   N. Kandasamy, R. Badrinarayanan, V. Kanamarlapudi, K. Tseng, B.-H. Soong, Performance Analysis of Machine-Learning Approaches for Modeling the Charging/Discharging Profiles of Stationary Battery Systems with Non-Uniform Cell Aging, Batteries. 3 (2017) 18. https://doi.org/10.3390/batteries3020018.

[19]   S. Wan, H. Jiang, Z. Guo, C. He, X. Liang, N. Djilali, T. Zhao, Machine learning-assisted design of flow fields for redox flow batteries, Energy Environ Sci. 15 (2022). https://doi.org/10.1039/d1ee03224k.

[20]   N. Ra, A. Bhattacharjee, Prediction of vanadium redox flow battery storage system power loss under different operating conditions: Machine learning based approach, Int J Energy Res. 46 (2022). https://doi.org/10.1002/er.8757.

[21]   A.A. Howard, T. Yu, W. Wang, A.M. Tartakovsky, Physics-informed CoKriging model of a redox flow battery, J Power Sources. 542 (2022). https://doi.org/10.1016/j.jpowsour.2022.231668.

[22]   X. Zhao., S. Jung, Long-term capacity fade forecasting of vanadium redox flow battery with Gaussian process regression combined with informer model, J Power Sources, 585 (2023). https://doi.org/10.1016/j.jpowsour.2023.233670

[23]   Q. Zheng, H. Zhang, F. Xing, X. Ma, X. Li, G. Ning, A three-dimensional model for thermal analysis in a vanadium flow battery, Appl Energy. 113 (2014). https://doi.org/10.1016/j.apenergy.2013.09.021.

[24]   H. Al-Fetlawi, A.A. Shah, F.C. Walsh, Non-isothermal modelling of the all-vanadium redox flow battery, Electrochim Acta. 55 (2009) 78–89. https://doi.org/10.1016/j.electacta.2009.08.009.

[25]   A. Tang, S. Ting, J. Bao, M. Skyllas-Kazacos, Thermal modelling and simulation of the all-vanadium redox flow battery, J Power Sources. 203 (2012). https://doi.org/10.1016/j.jpowsour.2011.11.079.

[26]   A. Tang, J. Bao, M. Skyllas-Kazacos, Thermal modelling of battery configuration and self-discharge reactions in vanadium redox flow battery, J Power Sources. 216 (2012). https://doi.org/10.1016/j.jpowsour.2012.06.052.

[27] Y. Yan, Y. Li, M. Skyllas-Kazacos, J. Bao, Modelling and simulation of thermal behaviour of vanadium redox flow battery, J Power Sources. 322 (2016). https://doi.org/10.1016/j.jpowsour.2016.05.011.

[28] C. Zhang, T.S. Zhao, Q. Xu, L. An, G. Zhao, Effects of operating temperature on the performance of vanadium redox flow batteries, Appl Energy. 155 (2015). https://doi.org/10.1016/j.apenergy.2015.06.002.

[29] A. Trovò, M. Guarnieri, Standby thermal management system for a kW-class vanadium redox flow battery, Energy Convers Manag. 226 (2020). https://doi.org/10.1016/j.enconman.2020.113510.

[30] A. Bhattacharjee, H. Saha, Development of an efficient thermal management system for Vanadium Redox Flow Battery under different charge-discharge conditions, Appl Energy. 230 (2018). https://doi.org/10.1016/j.apenergy.2018.09.056.

[31] Z. Wei, A. Bhattarai, C. Zou, S. Meng, T.M. Lim, M. Skyllas-Kazacos, Real-time monitoring of capacity loss for vanadium redox flow battery, J Power Sources. 390 (2018). https://doi.org/10.1016/j.jpowsour.2018.04.063.

[32] X. Su, X. Yan, C.-L. Tsai, Linear regression, Wiley Interdiscip Rev Comput Stat. 4 (2012) 275–294. https://doi.org/10.1002/wics.1198.

[33] D. Maulud, A.M. Abdulazeez, A Review on Linear Regression Comprehensive in Machine Learning, Journal of Applied Science and Technology Trends. 1 (2020). https://doi.org/10.38094/jastt1457.

[34] S. Weisberg, Applied Linear Regression: Third Edition, 2005. https://doi.org/10.1002/0471704091.

[35] A.I. Khuri, Introduction to Linear Regression Analysis, Fifth Edition by Douglas C. Montgomery, Elizabeth A. Peck, G. Geoffrey Vining, International Statistical Review. 81 (2013) 318–319. https://doi.org/10.1111/insr.12020_10.

[36] M. Awad, R. Khanna, Support Vector Regression, in Efficient Learning Machines, Apress, Berkeley, CA, 2015: pp. 67–80. https://doi.org/10.1007/978-1-4302-5990-9_4.

[37] B. Üstün, W.J. Melssen, L.M.C. Buydens, Visualisation and interpretation of Support Vector Regression models, Anal Chim Acta. 595 (2007) 299–309. https://doi.org/10.1016/j.aca.2007.03.023.

[38]  T. Chen, C. Guestrin, XGBoost: A scalable tree boosting system, in: Proceedings of the ACM SIGKDD International Conference on Knowledge Discovery and Data Mining, 2016. https://doi.org/10.1145/2939672.2939785.

[39]  T. Chen, T. He, A Scalable Tree Boosting, The Journal of Machine Learning Research. 21(20) (2020) 1–5.

[40]  A. Bhattacharjee, A. Roy, N. Banerjee, S. Patra, H. Saha, Precision dynamic equivalent circuit model of a Vanadium Redox Flow Battery and determination of circuit parameters for its optimal performance in renewable energy applications, J. Power Sources 396 (2018) 506–518, https://doi.org/10.1016/j.jpowsour.2018.06.017.

[41]  H. Wang, W L. Soong, S. A Pourmousavi, X. Zhang, N. Ertugrul, B. Xiong, Thermal dynamics assessment of vanadium redox flow batteries and thermal management by active temperature control, J Power Sources, 570 (2023) 233027, https://doi.org/10.1016/j.jpowsour.2023.233027

[42]  A. Bhattacharjee, H. Saha, Development of an efficient thermal management system for Vanadium Redox Flow Battery under different charge-discharge conditions, Appl Energy. 230 (2018) 1182-1192, https://doi.org/10.1016/j.apenergy.2018.09.056.

[43]  Q. He, Z. Li, D. Zhao, J. Yu, P. Tan, M. Guo, T. Liao, T. Zhao, M. Ni, A 3D modeling study on all vanadium redox flow battery at various operating temperatures, Energy. 282 (2023) 128934, https://doi.org/10.1016/j.energy.2023.128934.

[44]  Á. Cunha, F.P. Brito, J. Martins, N. Rodrigues, V. Monteiro, J L. Afonso, P. Ferreira, Assessment of the use of vanadium redox flow batteries for energy storage and fast charging of electric vehicles in gas stations, Energy, 115 (2) (2016) 1478-1494, https://doi.org/10.1016/j.energy.2016.02.118.